\documentclass[10pt,twocolumn,letterpaper]{article}

\usepackage[pagenumbers]{cvpr} 

\definecolor{cvprblue}{rgb}{0.21,0.49,0.74}
\usepackage[pagebackref,breaklinks,colorlinks,allcolors=cvprblue]{hyperref}

\def\confName{CVPR}
\def\confYear{2027}

\title{DiTraR: Diverse Scorers and Transcript-Guided Re-ranking for Video Corpus Moment Retrieval}

\author{Mohammad Alijanpour Shalmani\\
University of Central Florida\\
4000 Central Florida Blvd., Orlando, FL, USA\\
{\tt\small alijanpour@ucf.edu}
\and
Alale Rezvani Boroujeni\\
University of Central Florida\\
4000 Central Florida Blvd., Orlando, FL, USA\\
{\tt\small alale.rezvaniboroujeni@ucf.edu}
\and
Ali Amini\\
Kaunas University of Technology\\
K. Donelaičio St. 73, 44249 Kaunas, Lithuania\\
{\tt\small ali.amini@ktu.edu}
}

\begin{document}
\maketitle
\begin{abstract}
Video corpus moment retrieval asks a system to find, in a large collection of untrimmed videos with subtitles, the video and the time interval that a natural-language query describes. We find that most top-ranked errors of a strong corpus-level scorer come from choosing the wrong video, not from imprecise boundaries, and that re-weighting the evidence the scorer already uses does not fix them. We present DiTraR, which addresses video selection from two complementary directions. A witness-interval scorer decomposes the query into probes and scores every candidate interval of every corpus video exactly; a multi-scorer energy ensemble averages the energies of independently trained scorers before one exhaustive search; and a transcript verifier, a frozen language model that reads the time-stamped dialogue of the highest-ranked videos, re-orders the moments without changing their boundaries. With all settings frozen on development data and a single evaluation, DiTraR reaches 24.36 and 13.55 R@1 at IoU 0.5 and 0.7 on the TVR validation split. The ensemble improves queries about visual content, the verifier improves queries about dialogue, and their gains largely add up.
\end{abstract}

    
\section{Introduction}
\label{sec:intro}

Video corpus moment retrieval (VCMR) takes a natural-language query and returns the video, start time, and end time of a matching event from a collection of untrimmed videos~\cite{escorcia2019moments,lei2020tvr}. This capability lets a user search a video archive for a particular interaction even when the user knows neither the recording nor when the interaction occurred. The distinction matters in collections such as TV show Retrieval (TVR), where the same characters and settings recur across episodes and a few seconds of action or dialogue can determine which result answers the query~\cite{lei2020tvr}.

Existing methods approach this search at different temporal scales. Some encode videos and queries independently, allowing visual features to be computed offline, and learn similarities at the video and frame levels~\cite{zhang2021reloclnet}. Others use the query to fuse visual and subtitle features and localize a moment within shortlisted videos~\cite{hou2021conquer}. Training-free approaches also offer two routes to retrieval. RefCap indexes captions generated for candidate events~\cite{pan2025refcap}, while SCAC matches structured event chains and subsequently refines their temporal boundaries~\cite{zhao2026scac}.

A central difficulty is that evidence identifying an occurrence can persist well beyond the event itself. Consider the query \emph{a man in a black suit closes a door}. The suit may distinguish the man from another person who closes the same door, yet it remains visible before and after the door moves. Conversely, the closing motion gives a useful temporal cue, but it may recur elsewhere in the collection. A scorer that independently selects the strongest match for each phrase can therefore combine the suit from one shot with a door closing in another. Object-driven retrieval errors have been observed in VCMR~\cite{yoon2022squidnet}, and the VERIFIED benchmark shows how partially matching candidates make both static and dynamic details consequential~\cite{chen2024verified}. We therefore ask whether appearance can distinguish occurrences and action evidence can delimit them when both must support the same candidate.

We propose DiTraR, which keeps the full query, also decomposes it into probes for participants, objects and actions, and scores every candidate interval $(v,t_s,t_e)$ of every corpus video in one exhaustive search, so that a well-supported interval can change both the predicted timestamps and the selected video. Since most top-ranked errors come from choosing the wrong video, we improve video selection in two complementary ways: a multi-scorer energy ensemble that averages independently trained scorers, and a transcript verifier, a frozen language model that reads the dialogue of the highest-ranked videos and re-orders the moments without changing their boundaries. Our contributions are threefold: (i) an exact interval search in which query evidence must be witnessed within a common candidate interval; (ii) the ensemble and the verifier, which address visual and dialogue queries respectively and whose gains largely add up; and (iii) a controlled evaluation, with ensemble and verifier settings selected on development data, that reports end-to-end recall at rank one (R@1) at temporal IoU 0.5 and 0.7 on TVR~\cite{lei2020tvr}. On the validation split, DiTraR achieves 24.36 and 13.55 R@1 overall, and 19.82 and 10.77 on visual-only queries, respectively.

\section{Related Work}
\label{sec:related}

\paragraph{Text-to-video retrieval.}
Text-to-video retrieval ranks videos by relevance to a text query. Frozen in Time learns shared embeddings from captioned images and videos~\cite{bain2021frozen}, while CLIP4Clip aggregates pretrained frame embeddings~\cite{luo2022clip4clip}. Hierarchical Graph Reasoning matches events, actions, and entities at distinct semantic levels~\cite{chen2020hgr}, and Video-ColBERT scores token interactions across space and time~\cite{reddy2025videocolbert}. For queries describing only part of a video, MS-SL combines clip and frame similarities~\cite{dong2022prvr}, while DreamPRVR uses diffusion-guided registers to incorporate global context into local matching~\cite{li2026dreamprvr}. Video-level recall credits the correct video even when the supporting evidence falls outside the event.

\paragraph{Language-guided event localization.}
Temporal grounding locates a described event within a given video. CTRL scores clips and refines their boundaries~\cite{gao2017tall}. Moment-DETR jointly predicts intervals and clip saliency~\cite{lei2021momentdetr}. QD-DETR conditions video features on text early in the encoder~\cite{moon2023qddetr}, while Sim-DETR constrains interactions among overlapping predictions and aligns decoder queries with frames~\cite{tang2025simdetr}. DeCo decomposes descriptions into primitive events and uses masked reconstruction for novel verb--noun combinations~\cite{yang2023deco}. HieraMamba uses multiscale summaries to preserve boundary detail in long videos~\cite{an2026hieramamba}. Evaluating within a supplied video leaves cross-video score calibration untested.

\paragraph{Video corpus moment retrieval.}
VCMR couples video selection with temporal localization. The predicted interval must overlap the annotated event \emph{and} belong to the correct video. Escorcia et al.~\cite{escorcia2019moments} aligned language with regions in candidate moments across a collection, and TVR introduced a large-scale video--subtitle benchmark and the XML baseline~\cite{lei2020tvr}. ReLoCLNet uses video- and frame-level contrastive objectives to retain the efficiency of separate encoders~\cite{zhang2021reloclnet}. CONQUER instead makes multimodal fusion and localization query-dependent for shortlisted candidates~\cite{hou2021conquer}. SQuiDNet explicitly models the retrieval bias associated with query objects~\cite{yoon2022squidnet}, while EventFormer groups neighboring frames into events for corpus retrieval~\cite{hou2024eventformer}. VERIFIED tests whether these approaches can distinguish moments that satisfy different parts of a fine-grained description~\cite{chen2024verified}.

Recent work brings different sources of evidence to this decision. RefCap searches generated event captions~\cite{pan2025refcap}. SCAC constructs and compares query and video event chains without updating model parameters~\cite{zhao2026scac}. DeMUL models visual and subtitle evidence separately before combining them for localization~\cite{yang2026demul}. GenSpan uses decomposed subevents and generated motion sequences as temporal priors for multi-verb queries~\cite{sun2026genspan}, and VideoSearch-R1 refines its search query when verification of a retrieved video fails~\cite{lee2026videosearch}. Our comparison holds candidate intervals and frozen features fixed to isolate the effect of requiring appearance and action evidence to support the same occurrence.
\section{Method}
\label{sec:method}

\begin{figure*}[t]
  \centering
  \includegraphics[width=\linewidth]{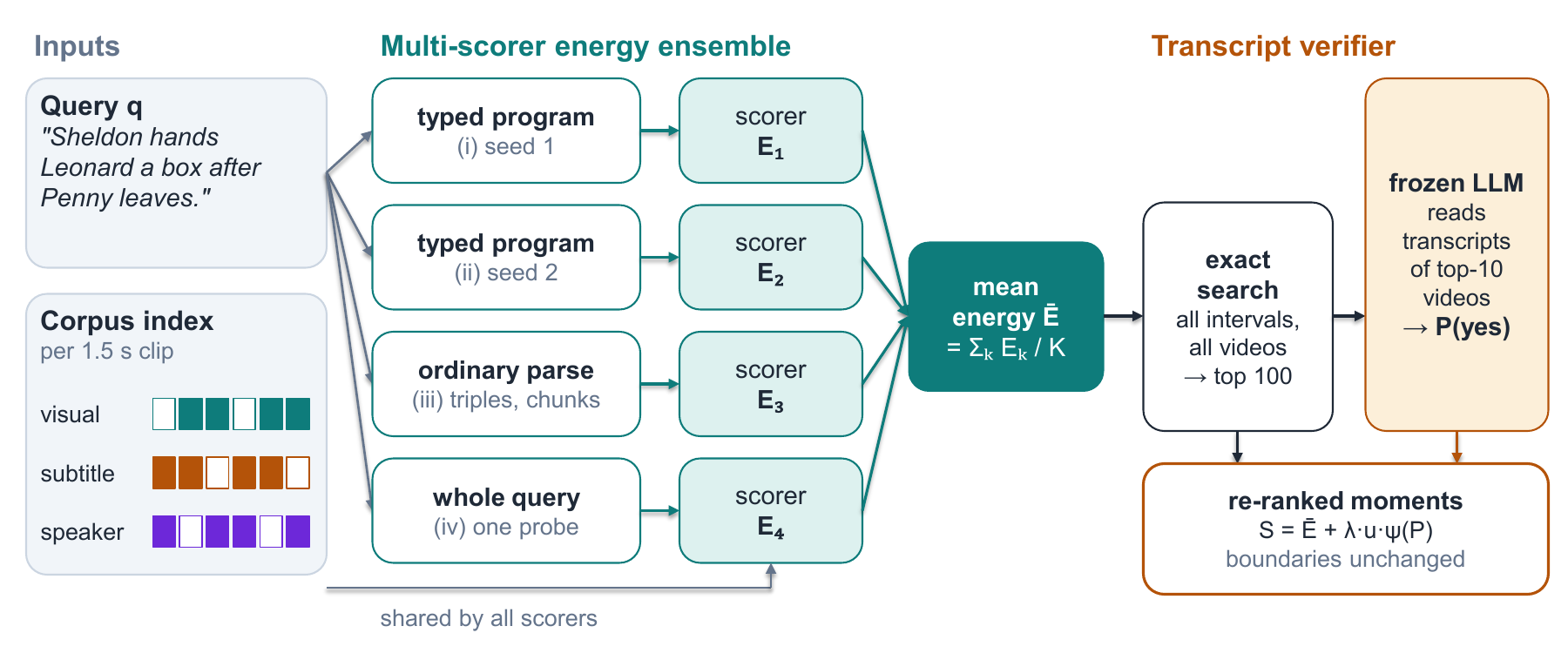}
  \caption{\textbf{Overview of DiTraR.} $K$ independently trained witness-interval scorers, which differ in query
  decomposition or random seed, assign an energy to every lattice interval of every corpus video. One exact search
  over the averaged energy returns the 100 best moments. A frozen language model reads the time-stamped transcripts of
  the highest-ranked videos, and its confidence that the moment occurs in each video re-orders the moments without
  changing their boundaries.}
  \label{fig:overview}
\end{figure*}

Given a natural-language query $q$ and a corpus $\mathcal{V}$ of untrimmed videos with time-stamped subtitles, video
corpus moment retrieval returns a ranked list of moments $(v, \tau^{s}, \tau^{e})$, where $v \in \mathcal{V}$ is a
video and $[\tau^{s}, \tau^{e}]$ an interval in it. A moment is correct only if both the video and the interval
match the annotation. DiTraR combines three components (\cref{fig:overview}). A \emph{witness-interval scorer}
decomposes the query into probes and assigns an energy to every candidate interval of every corpus video
(\cref{sec:method:scorer}). A \emph{multi-scorer energy ensemble} averages the energies of independently trained
scorers and searches the corpus once with the averaged energy (\cref{sec:method:ensemble}).
A \emph{transcript verifier}, a frozen language model that reads the dialogue of the highest-ranked videos,
re-ranks the resulting moments by how well each video's transcript supports the query (\cref{sec:method:verifier}).
The corpus is searched exhaustively; no component depends on the candidate list of an external retrieval model.

\subsection{Corpus representation}
\label{sec:method:index}

We split each video $v$ into $T_v$ clips of duration $\delta=1.5$\,s and precompute query-independent vectors $g_m(v,t)$ for $m\in\mathcal{M}=\{\mathrm{vis},\mathrm{sub},\mathrm{spk}\}$. A temporal encoder combines CLIP~\cite{radford2021clip}, ResNet~\cite{he2016resnet}, and SlowFast~\cite{feichtenhofer2019slowfast} features into the visual vector. A frozen sentence encoder~\cite{xiao2024cpack} embeds overlapping subtitle lines, and the speaker vector embeds their speaker names. Clips without dialogue use a learned \emph{no-speech} subtitle vector and a zero speaker vector. The boundary heads use $g_{\mathrm{bnd}}(v,t)$, which combines all three channels.

Candidate moments are the intervals of a lattice
$\Lambda_v = \{[s, s + \ell - 1] : 1 \le s \le T_v,\ \ell \in \mathcal{L},\ s + \ell - 1 \le T_v\}$
with $|\mathcal{L}| = 28$ lengths between 1 and 136 clips (1.5\,s to 204\,s): every length up to 16 clips, then
increasingly sparse lengths. On the validation split the lattice contains an interval with IoU $\ge 0.5$
($\ge 0.7$) to the annotated moment for 98.5\% (89.7\%) of the queries.

\subsection{Witness-interval scorer}
\label{sec:method:scorer}

\paragraph{Query decomposition.}
A decomposition operator $D$ maps the query to a set of probes
$D(q) = \{a_0, a_1, \dots, a_{A}\}$, where $a_0 = q$ is the whole query and $a_1, \dots, a_A$ ($A \le 8$) are
shorter phrases that each assert one fact about the moment. Our default operator builds a typed program on top of a
dependency parse~\cite{qi2020stanza}: event clauses with their arguments resolved (``Sheldon insults Raj'' from
``\dots after Sheldon insults him''), named participants, descriptions of people, objects and places, and reported
speech. Each probe carries its type and, from temporal connectives such as \emph{after}, \emph{before} and
\emph{while}, a prior on where it should be observed relative to the moment. Negated probes and the visual channel of
reported speech are masked. \Cref{sec:method:ensemble} introduces two further operators.

\paragraph{Gates and layouts.}
Each probe $a$ is embedded per channel, $f_m(a)$, with a text tower fine-tuned for that channel, and compared with
every clip by cosine similarity $\sigma_m(a, v, t) = \cos\big(f_m(a), g_m(v,t)\big)$. A gate
$\gamma_{a,m} \in [0, 1]$ decides how much channel $m$ should be consulted for probe $a$, and a layout distribution
$\rho_{a,m,\cdot}$ over $r \in \{\mathrm{in}, \mathrm{pre}, \mathrm{post}, \mathrm{video}\}$ decides where the
evidence should be read:
\begin{align}
  \gamma_{a,m} &= \mu_{a,m}\,\mathrm{sigm}\big(\pi_m(a) + h_{\gamma}(a, q)_m\big), \label{eq:gate}\\
  \rho_{a,m,\cdot} &= \mathrm{softmax}\big(\pi^{\rho}(a) + h_{\rho}(a, q)_m\big), \label{eq:layout}
\end{align}
where $\pi$ are fixed priors derived from the probe type, $h_{\gamma}$ and $h_{\rho}$ are small networks, and
$\mu_{a,m} \in \{0,1\}$ is the structural mask. No supervision other than the retrieval loss is used for the gates or
the layouts.

\begin{figure}[t]
  \centering
  \includegraphics[width=\linewidth]{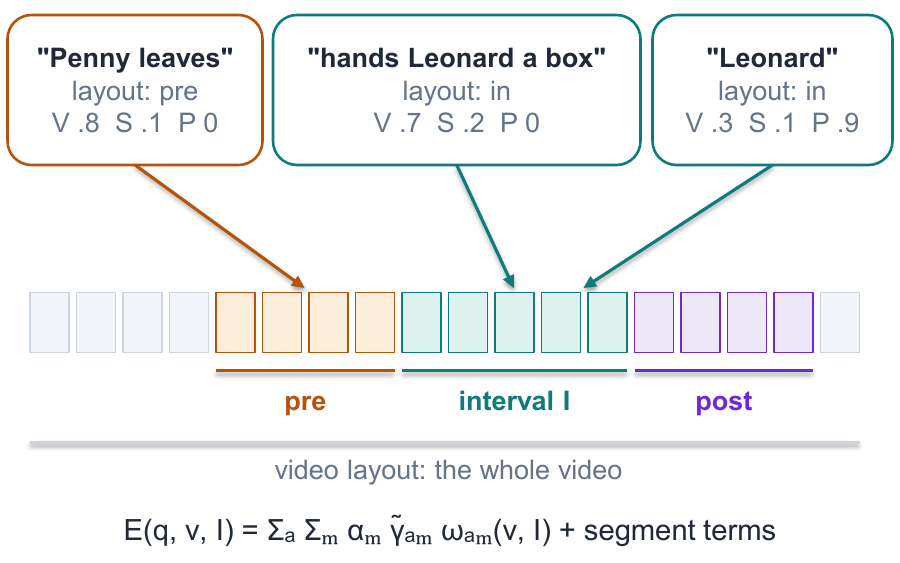}
  \caption{\textbf{Witnessed evidence for one candidate interval.} Each probe reads its similarity to the clips of a
  layout-specific window: inside the interval, just before it, just after it, or anywhere in the video. Gates decide
  which channels (visual, subtitle, speaker) each probe consults. The interval energy sums the gated evidence of all
  probes, which must be witnessed around one common interval.}
  \label{fig:witness}
\end{figure}

\paragraph{Witnessed evidence.}
For an interval $I = [s, e]$ let $W_{\mathrm{in}}(I) = I$, let $W_{\mathrm{pre}}(I)$ and $W_{\mathrm{post}}(I)$ be
the $w = 4$ clips immediately before and after $I$, and let $W_{\mathrm{video}}$ be the whole video. The evidence
that probe $a$ is witnessed for $I$ in channel $m$ is a soft maximum over each window, mixed over layouts inside the
logarithm:
\begin{equation}
  \omega_{a,m}(v, I) = \frac{1}{\beta_m}\log \sum_{r} \rho_{a,m,r}\,
  \frac{1}{|W_r(I)|}\sum_{t \in W_r(I)} e^{\beta_m \sigma_m(a, v, t)} .
  \label{eq:witness}
\end{equation}
\Cref{fig:witness} illustrates the windows. Channels add evidence about the same probe, while layouts are
alternative readings of where that evidence lies. With
the video layout alone, \cref{eq:witness} does not depend on $I$ and the score reduces to conventional late
interaction at the video level.

\paragraph{Interval energy.}
The energy of interval $I$ in video $v$ sums the gated evidence of all probes and a segment term:
\begin{equation}
  \begin{split}
  E(q, v, I) = {}& \sum_{a \in D(q)} \sum_{m \in \mathcal{M}} \alpha_m\, \tilde{\gamma}_{a,m}\, \omega_{a,m}(v, I)
     + \sum_{t \in I} \phi_q(v, t) \\
     &+ \log p^{\mathrm{s}}_q(s \mid v) + \log p^{\mathrm{e}}_q(e \mid v) + \lambda_q(e - s + 1),
  \end{split}
  \label{eq:energy}
\end{equation}
where $\phi_q(v,t)$ is a per-clip relevance logit, $p^{\mathrm{s}}_q$ and $p^{\mathrm{e}}_q$ are start and end
distributions within the video produced by a temporal convolution over the query--clip similarity curves, and
$\lambda_q$ is a learned length prior. Summing the evidence of dependent probes inflates the energy roughly in
proportion to the number of probes and lets training fit individual query--video pairs. We therefore cap the total
gate mass of a query per channel at a budget $B_m$:
\begin{equation}
  \tilde{\gamma}_{a,m} = \gamma_{a,m}\,\min\!\Big(1, \frac{B_m}{\sum_{a'} \gamma_{a',m}}\Big),
  \qquad B = (2.5,\ 1.6,\ 1.1).
  \label{eq:budget}
\end{equation}

\paragraph{Exact search.}
Every window mean in \cref{eq:witness} and every sum in \cref{eq:energy} is a difference of prefix sums along the
clip axis. One pass therefore scores all intervals of all videos exactly, at a cost of
$\mathcal{O}(|D(q)|\,|\mathcal{M}|\,T_v\,|\mathcal{L}|)$ per video, without proposals and without a first-stage
retriever. Each video contributes its 20 best intervals, the 600 best intervals of the corpus are refined by a
regression head that moves each boundary by a few seconds using $g_{\mathrm{bnd}}$, and temporal non-maximum
suppression at IoU 0.7 returns 100 moments. Since all energies live on one scale, the rank of a video is the rank of
its best interval; no separate video-retrieval score exists.

\paragraph{Training.}
For a batch of query--moment pairs drawn from $B_v = 64$ distinct videos, the energies of all lattice intervals of all
batch videos enter a single softmax,
\begin{equation}
  p_{\theta}(v, I \mid q) = \frac{\exp E_{\theta}(q, v, I)}
  {\sum_{v' \in \mathcal{B}}\sum_{I' \in \Lambda_{v'}} \exp E_{\theta}(q, v', I')},
  \label{eq:softmax}
\end{equation}
and the loss is the negative log of the probability mass on positive intervals (IoU $\ge 0.7$ with the annotation,
or the best lattice interval if none qualifies). Because the softmax runs over intervals of different videos, video
retrieval and localisation are learned with one objective. We add a symmetric moment-to-query term (weight 0.5), a
smooth-$\ell_1$ loss for the boundary refinement, and per-clip auxiliary terms. To provide hard negatives, 75\% of
the videos of a batch come from the same TV show.

\subsection{Multi-scorer energy ensemble}
\label{sec:method:ensemble}

A single scorer commits to one way of splitting the query and to one training run. We train $K$ scorers
$\{E_k\}_{k=1}^{K}$ independently with the same objective and data, but with different decomposition operators $D_k$
or random seeds, and search the corpus with their mean energy
\begin{equation}
  \bar{E}(q, v, I) = \frac{1}{K}\sum_{k=1}^{K} E_k(q, v, I).
  \label{eq:ensemble}
\end{equation}
Since every member defines a distribution $p_k(v, I \mid q) \propto \exp E_k(q, v, I)$ over the same corpus lattice,
\cref{eq:ensemble} corresponds to the normalised geometric mean
$\bar{p} \propto \prod_k p_k^{1/K}$, a product of experts~\cite{hinton2002poe}: an interval ranks highly only if all
members agree, which suppresses the false positives of individual members. All members are trained with the
corpus-level softmax of \cref{eq:softmax}, so their energies share a scale and are averaged with equal weights. The
refinement offsets are averaged likewise, and the search of \cref{sec:method:scorer} runs unchanged on $\bar{E}$.

We use $K = 4$ members:
(i) the typed program described above;
(ii) the same configuration with a different random seed;
(iii) an \emph{ordinary} decomposition into subject--verb--object triples and noun chunks read from the dependency
tree, without coreference, type priors or masks, so that gates and layouts are learned without parser priors; and
(iv) a \emph{holistic} scorer that uses the whole query as its only probe.
No member is much stronger than (i), and (iv) is clearly weaker, but each improves on (i) when the two are averaged
(\cref{sec:exp:ensemble}).

\subsection{Transcript verifier}
\label{sec:method:verifier}

The scorer encodes each subtitle line on its own, so it cannot follow a conversation, resolve who says what to whom,
or match a described situation to dialogue that paraphrases it. For such queries the correct video is often among the
first candidates but not at the top. The verifier therefore reads the complete transcript of each candidate video.

\paragraph{Candidate videos.}
For each query we collect the set $C_q$ of the $n = 10$ highest-ranked distinct videos of the ensemble and of the
first member alone (13.1 videos per query on average on validation). The verifier never sees videos outside $C_q$ and never adds
candidates.

\paragraph{Reading.}
Each video is rendered as a script that lists its duration and every subtitle line with its time stamps and speaker
name. A frozen instruction-tuned language model~\cite{qwen35} receives the script and a numbered list of up to
$G = 8$ queries that have this video in their candidate set, and answers for each query whether the described moment
occurs in the video. Grouping queries per transcript amortises the cost of reading it. Decoding is greedy and
constrained by a regular expression to one line per query of the form \texttt{i: yes|no start end}. From the token
log-probabilities at the decision position we obtain
\begin{equation}
  p_{q,v} = \frac{\exp \ell_{\mathrm{yes}}}{\exp \ell_{\mathrm{yes}} + \exp \ell_{\mathrm{no}}}.
  \label{eq:pyes}
\end{equation}
The span that the model writes is not used.

\paragraph{Re-ranking.}
The moments returned by the search are re-ordered by
\begin{equation}
  S(q, v, I) = \bar{E}(q, v, I) + \lambda\, u\, \psi_{q,v},
  \label{eq:rerank}
\end{equation}
where $\psi_{q,v} = \mathrm{clip}\big(\mathrm{logit}\, p_{q,v}, -c, c\big)$ for $v \in C_q$ and $\psi_{q,v} = -c$
otherwise,
with $c = 4$. Here $u$ is a label-free estimate of the energy scale of the corpus search, the median over queries of
the standard deviation of the energies of the intervals retained in the top-ranked video, so that $\lambda$ is
expressed in units of typical within-video energy differences. The single hyper-parameter $\lambda$ is chosen on
development data. All moments of one video move by the same amount, so the verifier changes which video is ranked first
but neither the boundaries of any moment nor the set of returned moments; recall at 100 is unchanged by
construction.

\section{Experiments}
\label{sec:exp}

\subsection{Experimental setup}
\label{sec:exp:setup}

\paragraph{Dataset.}
TVR~\cite{lei2020tvr} contains 109K queries over 21.8K videos from six TV shows with time-stamped, speaker-labeled subtitles. Training has 87,175 queries over 17,435 videos. Validation has 10,895 queries over 2,179 videos. We use modality labels (74.3\% visual, 8.8\% subtitle, 16.8\% both on validation) solely to report subgroup results. Since test annotations are private, we follow prior work~\cite{yang2026demul,sun2026genspan,hou2024prem} and report official validation results.

\paragraph{Development protocol.}
We hold out whole training episodes for \emph{Dev-Tune} (5,385 queries, 1,077 videos) and \emph{Dev-Test} (5,170 queries, 1,034 videos), training on the remaining 76,620 queries. We select settings on Dev-Tune and evaluate once on Dev-Test. Both corpora are roughly half the validation corpus size, limiting direct score comparisons. Validation informed single-scorer development. We froze the ensemble and verifier on Dev-Tune before evaluating either on validation, retrained each member on the full training split, and evaluated the complete system once.

\paragraph{Metrics.}
Following~\cite{lei2020tvr}, R@$k$ measures the fraction of queries with a top-$k$ moment in the correct video meeting the temporal IoU threshold. We report $k\in\{1,5,10,100\}$ at IoU 0.5 and 0.7, with R@1 primary at both thresholds. VR@1 measures how often the top moment belongs to the correct video. Paired differences use 95\% bootstrap intervals from 2,000 video-level resamples to account for correlated queries~\cite{efron1993bootstrap}.

\paragraph{Implementation details.}
We combine CLIP ViT-B/32 frame features~\cite{radford2021clip} with TVR ResNet-152/SlowFast features~\cite{he2016resnet,feichtenhofer2019slowfast} (1,024-D after PCA). Frozen bge-base embeds subtitles~\cite{xiao2024cpack}. Fine-tuned CLIP and bge encode visual and subtitle/speaker probes, respectively. Stanza provides dependency parses~\cite{qi2020stanza}. Each 193M-parameter scorer takes ${\sim}1$ H100 GPU-hour for 8 AdamW epochs (64 videos/batch, 0.01 weight decay, 300 warm-up steps). Learning rates are $1.5\times10^{-5}$ for text encoders and $3\times10^{-4}$ elsewhere. vLLM~\cite{kwon2023vllm} serves Qwen3.5-27B~\cite{qwen35} in bfloat16 (4,096-token context, prefix caching and thinking disabled). Tuning only verifier weight and ensemble size by mean R@1 across both Dev-Tune thresholds selected $\lambda=1.0$ from $\{0.25,0.5,1.0,2.0\}$ and four members from $\{3,4\}$.

\begin{table*}[t]
  \centering
  \caption{\textbf{Comparison with the state of the art on the TVR validation split} (10,895 queries, 2,179 videos).
  \emph{Candidates}: where the videos in which moments are localised come from; HERO top-10 means that moments are
  ranked only inside the ten videos returned by HERO~\cite{li2020hero} (HERO$^{*}$: HERO re-trained by the authors).
  RN: ResNet, SF: SlowFast.
  \emph{Extra models}: models beyond the retrieval network that are used at inference. ``--'': not reported.
  $^{\dagger}$Reported by~\cite{hou2024prem}; $^{\ddagger}$reported by~\cite{sun2026genspan}, whose IoU 0.7 values for
  these methods match the original papers. Our rows are single runs.}
  \label{tab:sota}
  \footnotesize
  \setlength{\tabcolsep}{4.2pt}
  \begin{tabular}{@{}llllccccc@{}}
    \toprule
    & & & & \multicolumn{2}{c}{R@1} & \multicolumn{2}{c}{R@10} & R@100 \\
    \cmidrule(lr){5-6}\cmidrule(lr){7-8}\cmidrule(l){9-9}
    Method & Visual features & Candidates & Extra models & 0.5 & 0.7 & 0.5 & 0.7 & 0.7 \\
    \midrule
    XML~\cite{lei2020tvr}               & RN+I3D & own (top-100) & --                 & 5.28  & 2.62  & 15.90 & 9.05  & 22.47 \\
    HAMMER~\cite{zhang2020hammer}       & RN+I3D & own           & --                 & 9.19  & 5.13  & 21.28 & 11.38 & 16.71 \\
    ReLoCLNet~\cite{zhang2021reloclnet} & RN+I3D & own           & --                 & 8.03  & 4.15  & 21.37 & 14.06 & 32.42 \\
    HERO~\cite{li2020hero}              & RN+SF & own          & --                 & --    & 5.13  & --    & 16.26 & 24.55 \\
    CONQUER~\cite{hou2021conquer}       & RN+SF & HERO top-10  & HERO               & 13.57$^{\dagger}$ & 7.76 & -- & 22.49 & 35.17 \\
    SQuiDNet~\cite{yoon2022squidnet}    & RN+SF & own top-10   & --                 & 15.05$^{\ddagger}$ & 8.52  & 33.20$^{\ddagger}$ & 28.41$^{\ddagger}$ & --    \\
    MINUTE~\cite{hou2023minute}         & RN+SF & HERO$^{*}$ top-10 & HERO$^{*}$     & 19.22 & 10.70 & --    & 29.37 & 45.09 \\
    EventFormer~\cite{hou2024eventformer} & RN+SF & own top-10 & --                 & 16.04$^{\ddagger}$ & 10.12 & 35.12$^{\ddagger}$ & 27.54 & 42.88 \\
    PREM~\cite{hou2024prem}             & RN+SF & own top-10   & --                 & 16.08 & 9.01  & --    & --    & --    \\
    PREM~\cite{hou2024prem}             & RN+SF & HERO$^{*}$ top-10 & HERO$^{*}$     & 19.74 & 11.12 & --    & --    & --    \\
    GenSpan~\cite{sun2026genspan}       & CLIP           & not stated    & LLaMA-3.1, CogVideoX & 18.77 & 12.12 & 40.56 & 33.31 & -- \\
    DeMUL~\cite{yang2026demul}          & RN+SF & HERO top-10  & HERO               & 21.37 & 11.92 & 39.31 & 31.34 & 39.79 \\
    \midrule
    Single scorer (ours)                & CLIP+RN+SF & all videos & --             & 20.23 & 11.07 & 44.30 & 31.40 & 47.97 \\
    Single scorer + verifier (ours)     & CLIP+RN+SF & all videos & Qwen3.5-27B    & 22.58 & 12.25 & 46.87 & 34.05 & 47.97 \\
    DiTraR w/o verifier                  & CLIP+RN+SF & all videos & --             & 22.83 & 12.76 & 47.81 & 34.96 & 51.69 \\
    DiTraR                               & CLIP+RN+SF & all videos & Qwen3.5-27B    & \textbf{24.36} & \textbf{13.55} & \textbf{49.36} & \textbf{36.70} & \textbf{51.69} \\
    \bottomrule
  \end{tabular}
\end{table*}

\subsection{Comparison with the state of the art}
\label{sec:exp:sota}

\Cref{tab:sota} compares DiTraR with published results on the validation split. The single witness-interval scorer
(20.23 / 11.07 R@1 at IoU 0.5 / 0.7) is the strongest method at IoU 0.5 that does not localise inside
another model's candidate list, but it remains below DeMUL~\cite{yang2026demul}, which ranks moments inside HERO's top
ten videos. The ensemble raises R@1 to 22.83 / 12.76, and the transcript verifier to 24.36 /
13.55, a paired gain of $+4.13$ [3.48, 4.76]\,pp and $+2.48$ [1.97, 2.99]\,pp over the single scorer. DiTraR exceeds
the best published R@1 at both thresholds (DeMUL, 21.37, at IoU 0.5; GenSpan, 12.12, at IoU 0.7), and since it
searches all 2,179 videos the margin grows deeper in the list (R@100 at IoU 0.7: 51.69 against 39.79). Two
differences in setting matter: we use CLIP frame features in addition to ResNet and SlowFast, and the verifier relies
on a 27B-parameter language model (\cref{sec:exp:cost}). GenSpan~\cite{sun2026genspan} also uses a language model,
but to select subtitle lines and decompose the query for its localiser, not to rank videos.

\begin{table*}[t]
  \centering
  \caption{\textbf{Contribution of each component.} Recall at IoU 0.5 and 0.7, and VR@1, for every combination of the
  two components. The verifier only re-orders the returned moments, so it leaves R@100 unchanged; the ensemble changes
  which moments are returned. Best value per column in bold.}
  \label{tab:components}
  \footnotesize
  \setlength{\tabcolsep}{3.4pt}
  \begin{tabular}{@{}cc ccccccc ccccccc@{}}
    \toprule
    & & \multicolumn{7}{c}{Dev-Test (5,170 queries)} & \multicolumn{7}{c}{Validation (10,895 queries)} \\
    \cmidrule(lr){3-9}\cmidrule(l){10-16}
    & & \multicolumn{2}{c}{R@1} & \multicolumn{2}{c}{R@10} & \multicolumn{2}{c}{R@100} & &
        \multicolumn{2}{c}{R@1} & \multicolumn{2}{c}{R@10} & \multicolumn{2}{c}{R@100} & \\
    \cmidrule(lr){3-4}\cmidrule(lr){5-6}\cmidrule(lr){7-8}\cmidrule(lr){10-11}\cmidrule(lr){12-13}\cmidrule(lr){14-15}
    Ens. & Ver. & 0.5 & 0.7 & 0.5 & 0.7 & 0.5 & 0.7 & VR@1 & 0.5 & 0.7 & 0.5 & 0.7 & 0.5 & 0.7 & VR@1 \\
    \midrule
               &            & 19.98 & 10.85 & 44.56 & 33.02 & 66.91 & 51.10 & 41.86
                            & 20.23 & 11.07 & 44.30 & 31.40 & 63.26 & 47.97 & 42.47 \\
               & \checkmark & 22.05 & 11.88 & 46.79 & 34.95 & 66.91 & 51.10 & 48.24
                            & 22.58 & 12.25 & 46.87 & 34.05 & 63.26 & 47.97 & 49.76 \\
    \checkmark &            & 22.32 & 12.50 & 48.55 & 35.82 & \textbf{69.52} & \textbf{53.23} & 44.45
                            & 22.83 & 12.76 & 47.81 & 34.96 & \textbf{66.69} & \textbf{51.69} & 45.56 \\
    \checkmark & \checkmark & \textbf{23.98} & \textbf{13.29} & \textbf{50.02} & \textbf{37.08} & \textbf{69.52} & \textbf{53.23} & \textbf{49.69}
                            & \textbf{24.36} & \textbf{13.55} & \textbf{49.36} & \textbf{36.70} & \textbf{66.69} & \textbf{51.69} & \textbf{51.49} \\
    \bottomrule
  \end{tabular}
\end{table*}

\begin{table}[t]
  \centering
  \caption{\textbf{Paired differences} in R@1 (percentage points) with 95\% video-cluster bootstrap intervals.}
  \label{tab:paired}
  \footnotesize
  \setlength{\tabcolsep}{2.6pt}
  \begin{tabular}{@{}llcc@{}}
    \toprule
    Comparison & Split & IoU 0.5 & IoU 0.7 \\
    \midrule
    Ens.\ $-$ single  & Dev-Test & $+$2.34 {\scriptsize[1.49, 3.15]} & $+$1.64 {\scriptsize[0.91, 2.40]} \\
                      & Val.  & $+2.60$ {\scriptsize[2.00, 3.17]} & $+1.69$ {\scriptsize[1.22, 2.16]} \\
    Ver.\ $-$ single  & Dev-Test & $+$2.07 {\scriptsize[1.45, 2.69]} & $+$1.03 {\scriptsize[0.54, 1.47]} \\
                      & Val.  & $+2.35$ {\scriptsize[1.90, 2.76]} & $+1.18$ {\scriptsize[0.86, 1.51]} \\
    DiTraR $-$ single  & Dev-Test & $+$4.00 {\scriptsize[3.08, 4.91]} & $+$2.44 {\scriptsize[1.66, 3.23]} \\
                      & Val.  & $\mathbf{+4.13}$\ {\scriptsize[3.48, 4.76]} & $\mathbf{+2.48}$\ {\scriptsize[1.97, 2.99]} \\
    DiTraR $-$ Ens.    & Dev-Test & $+$1.66 {\scriptsize[1.06, 2.28]} & $+$0.79 {\scriptsize[0.33, 1.28]} \\
                      & Val.  & $+1.53$ {\scriptsize[1.07, 1.98]} & $+0.79$ {\scriptsize[0.45, 1.13]} \\
    \midrule
    \multicolumn{4}{@{}l}{\emph{Visual-only queries} ($n = 3{,}797$ and $8{,}097$)} \\
    DiTraR $-$ single  & Dev-Test & $+$3.19 {\scriptsize[2.21, 4.14]} & $+$2.08 {\scriptsize[1.29, 2.92]} \\
                      & Val.  & $+$2.68 {\scriptsize[1.98, 3.39]} & $+$1.56 {\scriptsize[1.01, 2.13]} \\
    \bottomrule
  \end{tabular}
\end{table}

\subsection{Ablation studies}
\label{sec:exp:ablation}

\paragraph{Both components contribute, and their gains largely add up.}
On validation (\cref{tab:components,tab:paired}) the ensemble improves R@1 by $+2.60$ / $+1.69$ points and the
verifier by $+2.35$ / $+1.18$; together they give $+4.13$ / $+2.48$, most of the sum of the two separate gains,
because both partly repair the same queries. Dev-Test shows the same pattern. Every paired interval excludes
zero at both thresholds on Dev-Test and on validation, including the gain of the verifier on top of the ensemble
($+1.53$ / $+0.79$ on validation). On visual-only queries, three quarters of the benchmark, the full model gains
$+2.68$ / $+1.56$ on validation.

\begin{table}[t]
  \centering
  \caption{\textbf{Ensemble composition} (R@1 at IoU 0.5). \emph{Alone}: each member by itself. $\Delta$: energy
  average of the listed members, as the change over member (i) alone. Selected intervals and IoU 0.7 values are
  given in the text.}
  \label{tab:ensemble}
  \footnotesize
  \setlength{\tabcolsep}{4pt}
  \begin{tabular}{@{}lcccc@{}}
    \toprule
    & \multicolumn{2}{c}{Alone} & \multicolumn{2}{c}{$\Delta$ with (i)} \\
    \cmidrule(lr){2-3}\cmidrule(l){4-5}
    Member & Dev-Test & Val. & Dev-Test & Val. \\
    \midrule
    (i) typed program      & 19.98 & 20.23 & --      & --      \\
    (ii) typed, new seed   & 19.25 & 20.43 & $+$0.81 & $+$2.02 \\
    (iii) ordinary parse   & 20.29 & 20.56 & $+$1.90 & $+$1.62 \\
    (iv) whole query       & 18.47 & 18.10 & $+$1.37 & $+$1.14 \\
    \midrule
    (i)+(ii)+(iii)         & 22.26 & 22.69 & $+$2.28 & $+$2.46 \\
    (i)--(iv) (DiTraR)      & \textbf{22.32} & \textbf{22.83} & $+$\textbf{2.34} & $+$\textbf{2.60} \\
    \bottomrule
  \end{tabular}
\end{table}

\paragraph{Independently trained scorers make different errors.}
\label{sec:exp:ensemble}
\Cref{tab:ensemble} adds the members one at a time. Averaging the default scorer with any other member improves R@1
on both splits, although no member alone beats the default by more than 0.4 points; even the whole-query scorer, 2.1
points weaker alone on validation, adds $+1.14$ $[0.58, 1.68]$ as a partner. Which kind of diversity helps most is not
consistent: on Dev-Test a different decomposition adds more than a second seed ($+1.90$ against $+0.81$), whereas on
validation the second seed adds the most ($+2.02$ $[1.48, 2.54]$ against $+1.62$ $[1.06, 2.17]$ for the ordinary
parse). We therefore attribute the gain to averaging independently trained scorers, not to decomposition diversity in
particular. At IoU 0.7 the full ensemble adds $+1.64$ on Dev-Test and $+1.69$ on validation. The gains saturate (the
fourth member adds $+0.14$ on validation), and unlike the verifier the ensemble also lifts the deeper ranks (R@100 at
IoU 0.5 from 63.26 to 66.69).

\begin{table}[t]
  \centering
  \caption{\textbf{Verifier weight} $\lambda$ on Dev-Tune, the only values tried. The selected value (bold) is used
  everywhere else.}
  \label{tab:lambda}
  \footnotesize
  \setlength{\tabcolsep}{4pt}
  \begin{tabular}{@{}ccccc@{}}
    \toprule
    & \multicolumn{2}{c}{Single + verifier} & \multicolumn{2}{c}{DiTraR} \\
    \cmidrule(lr){2-3}\cmidrule(l){4-5}
    $\lambda$ & R@1 (0.5 / 0.7) & VR@1 & R@1 (0.5 / 0.7) & VR@1 \\
    \midrule
    0.25          & 20.45 / 11.05 & 44.49 & 22.58 / 12.72 & 46.93 \\
    0.5           & 21.34 / 11.48 & 47.04 & 23.27 / 12.96 & 48.97 \\
    \textbf{1.0}  & 21.69 / 11.51 & 48.49 & \textbf{23.42} / \textbf{13.02} & \textbf{49.97} \\
    2.0           & 21.34 / 11.42 & 48.43 & 22.90 / 12.72 & 49.66 \\
    \bottomrule
  \end{tabular}
\end{table}

\paragraph{The verifier is robust to its weight.}
\Cref{tab:lambda} lists the four weights we tried. Every value improves both systems, and over an eightfold range of
$\lambda$ the R@1 of DiTraR at IoU 0.5 varies by less than one point (22.58 to 23.42). At $\lambda = 2$ the verifier
starts to override confident energy differences: R@1 decreases slightly while VR@1 stays high.

\begin{table}[t]
  \centering
  \caption{\textbf{R@1 by annotated query type} on validation (IoU 0.5 / 0.7). The labels are never used by the
  model.}
  \label{tab:types}
  \footnotesize
  \setlength{\tabcolsep}{3.5pt}
  \begin{tabular}{@{}lccc@{}}
    \toprule
    & Video & Subtitle & Both \\
    System & ($n$\,=\,8,097) & ($n$\,=\,964) & ($n$\,=\,1,834) \\
    \midrule
    Single scorer   & 17.14 / \phantom{0}9.21 & 25.93 / 14.00 & 30.86 / 17.72 \\
    + verifier      & 18.27 / \phantom{0}9.86 & 33.20 / 17.53 & 36.04 / 20.07 \\
    + ensemble      & 19.65 / 10.63 & 27.59 / 15.87 & 34.35 / 20.50 \\
    DiTraR           & \textbf{19.82} / \textbf{10.77} & \textbf{33.82} / \textbf{18.67} & \textbf{39.42} / \textbf{23.12} \\
    \bottomrule
  \end{tabular}
\end{table}

\begin{figure}[t]
  \centering
  \includegraphics[width=\linewidth]{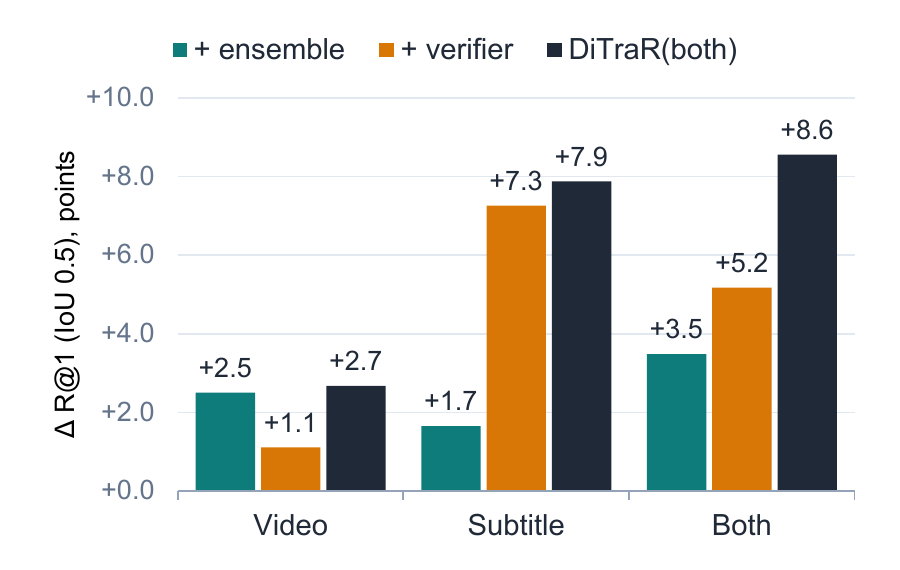}
  \caption{\textbf{Gains by query type} on validation (R@1 at IoU 0.5, change over the single scorer). The ensemble
  helps most on visual and mixed queries; the transcript verifier helps mainly on queries that refer to the
  dialogue.}
  \label{fig:types}
\end{figure}

\paragraph{The two components address different queries.}
\Cref{tab:types,fig:types} break the validation results down by annotated query type. On visual queries almost all of
the gain comes from the ensemble ($+2.51$ at IoU 0.5); on top of it the verifier, which reads only dialogue, adds
$+0.17$ $[-0.33, 0.66]$. On subtitle and mixed queries the verifier adds $+6.22$ $[4.53, 7.91]$ and $+5.07$
$[3.86, 6.33]$ over the ensemble, and DiTraR ranks the correct video first for 68.8\% and 71.9\% of them, against
50.4\% and 56.8\% for the single scorer. The ensemble improves evidence the scorer already uses, while the verifier
adds reasoning over whole conversations, which the clip-wise subtitle encoding cannot provide. Dev-Test shows the same
split.

\begin{table}[t]
  \centering
  \caption{\textbf{Design of the single scorer} (R@1 at IoU 0.5 / 0.7). Each row is a separate training run;
  validation runs are trained on the full training split. Rows without the budget are trained without
  \cref{eq:budget}. Runs of one configuration differ by up to 0.4 points: the default scores 20.21 / 10.70 (Dev-Test) and
  20.60 / 11.52 (validation) here, and 19.98 / 10.85 and 20.23 / 11.07 in the run used elsewhere.}
  \label{tab:scorer}
  \footnotesize
  \setlength{\tabcolsep}{4pt}
  \begin{tabular}{@{}lcc@{}}
    \toprule
    Variant & Dev-Test & Validation \\
    \midrule
    Whole query only                    & 18.05 / 10.37 & 17.87 / 10.11 \\
    Probes, video-level matching        & 17.14 / \phantom{0}8.94 & -- \\
    Probes, common interval             & 20.68 / 11.39 & 18.24 / \phantom{0}9.78 \\
    \quad + budget (default)            & 20.21 / 10.70 & \textbf{20.60} / \textbf{11.52} \\
    \quad + budget, ordinary parse      & 20.48 / 10.89 & -- \\
    \bottomrule
  \end{tabular}
\end{table}

\paragraph{Design of the single scorer.}
\Cref{tab:scorer} ablates the scorer itself. Matching several probes helps only if they must be witnessed in a common
interval: the same probes matched at the video level fall below the whole-query scorer (17.14 against 18.05 on Dev-Test),
whereas the common-interval constraint adds 2.6 points. Without the evidence budget of \cref{eq:budget}, this gain does
not carry over from development to validation. The run trained on the full split fits the training queries much more
closely (final training loss 0.81 against 1.96 on the development split) and gains only $+0.37$ over the
whole-query scorer on validation. With the budget the gain on validation is $+2.73$ $[1.96, 3.42]$. The typed parse
itself is not the source of the improvement: with the same budget, an ordinary parse performs as well
($-0.27$ $[-1.26, 0.68]$ for typed minus ordinary on Dev-Test; 20.23 against 20.56 on validation for the members of
\cref{tab:ensemble}).

\begin{table}[t]
  \centering
  \caption{\textbf{Training-free re-scoring with the scorer's own features} on Dev-Test. Each rule re-ranks a shared pool
  of candidates from the 100 best videos and was tuned on Dev-Tune with at most four configurations.}
  \label{tab:trainingfree}
  \footnotesize
  \setlength{\tabcolsep}{4pt}
  \begin{tabular}{@{}lcc@{}}
    \toprule
    Re-scoring rule & R@1 (0.5 / 0.7) & $\Delta$ (0.5) \\
    \midrule
    None (single scorer)                   & 19.98 / 10.85 & -- \\
    Whole-query visual contrast            & 19.75 / 10.89 & $-$0.23 \\
    Target / support / order evidence      & 19.88 / 10.72 & $-$0.10 \\
    Dialogue-aware switch                  & 19.85 / 10.54 & $-$0.13 \\
    Stated order of two actions            & 19.98 / 10.87 & \phantom{$-$}0.00 \\
    \midrule
    Ensemble (learned)                     & 22.32 / 12.50 & $+$2.34 \\
    Transcript verifier                    & 22.05 / 11.88 & $+$2.07 \\
    \bottomrule
  \end{tabular}
\end{table}

\paragraph{Re-scoring with the same features does not improve R@1.}
\Cref{tab:trainingfree} evaluates four rules using the scorer's clip embeddings, all tuned on Dev-Tune: (i) contrast full-query scores inside and around the candidate interval, (ii) score action, participant, and temporal cues separately, (iii) use subtitles to switch between the original ranking and rule (ii), and (iv) reward the stated action order. Rule (iv) applies to only 3.1\% of queries. With one vector per 1.5\,s clip, stated and reversed orders score nearly identically (0.696 vs.\ 0.694). These results favor scorers with complementary errors or sources of evidence beyond the clip embeddings.

\begin{figure}[t]
  \centering
  \includegraphics[width=\linewidth]{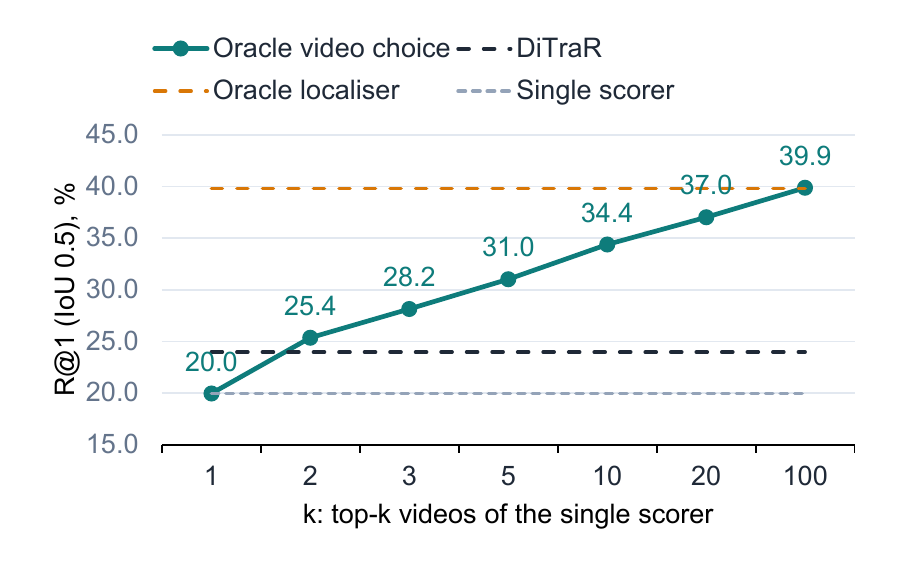}
  \caption{\textbf{Where the remaining errors are} (Dev-Test, R@1 at IoU 0.5). An oracle that picks the correct video
  whenever it is among the top-$k$ videos of the single scorer, and keeps the scorer's best moment in it, bounds what
  video selection alone can achieve. Horizontal lines: the single scorer, DiTraR, and an oracle localiser applied to the
  single scorer's top video.}
  \label{fig:headroom}
\end{figure}

\subsection{Analysis}
\label{sec:exp:analysis}

\paragraph{Gains from video selection.}
On validation, DiTraR raises VR@1 from 42.47\% to 51.49\%, while localisation given the correct video (R@1/VR@1 at IoU 0.5) stays near 47\% (47.6\% vs.\ 47.3\%). The ensemble alone improves both measures (45.56\%, 50.1\%). The verifier alone achieves higher VR@1 (49.76\%) but lower conditional localisation (45.4\%). Dev-Test shows the same pattern.

\paragraph{Headroom.}
In \Cref{fig:headroom}, oracle video selection among the single scorer's top 2 or 10 videos yields 25.4\% or 34.4\% R@1 at IoU 0.5, respectively. DiTraR approaches the top-2 oracle at 24.0\%. Oracle localisation within the scorer's top video yields 39.8\% at IoU 0.5 and 35.1\% at IoU 0.7, suggesting greater headroom for localisation, especially at stricter IoU.

\paragraph{Cost.}
\label{sec:exp:cost}
Each scorer indexes clips at 4.8\,kB per clip and searches 2,179 validation videos in 17\,ms per query on one H100. Four members take under 0.1\,s per query and cost four times as much to train. Verification dominates inference at 0.48\,s per query offline across 142,173 query--video pairs on one H100. It can rerank any system's candidates using the existing index.

\paragraph{Limitations.}
On visual queries (74\% of TVR), the dialogue-only verifier adds 1.12 points to the single scorer and 0.17 to the ensemble (95\% interval $[-0.33, 0.66]$). Pretraining may have included TVR transcripts. We provide the verifier with no query annotations or moment labels. Each member was trained once, so bootstrap intervals capture sampling uncertainty only. Repeated runs of one configuration differ by about 0.4 points (\cref{tab:scorer}). Private test annotations limit comparisons with prior work to validation.

\section{Conclusion}
\label{sec:conclusion}

DiTraR averages the energies of independently trained witness-interval scorers to search every candidate interval in every video exactly. A frozen language model verifies the top-ranked videos against their transcripts. With settings fixed on development data, DiTraR achieves 24.36 and 13.55 R@1 on TVR validation at IoU 0.5 and 0.7, respectively, improving on a single scorer by 4.13 and 2.48 points. The ensemble helps visual and mixed queries, while verification helps dialogue queries. Reweighting the scorer's own evidence yields no gain.

Visual queries account for three quarters of the benchmark and benefit mainly from the ensemble, motivating a verifier that also inspects frames. In-video localisation remains a priority. DiTraR ranks the correct video first for half of all queries but correctly localises fewer than half of those moments.

\section*{LLM Usage Statement}
We used ChatGPT to identify relevant literature, help draft the manuscript, edit for grammar, clarity, and concision. We also used Claude Code to generate portions of the experimental code. The authors take full responsibility for the manuscript, implementation, and reported results.
\newcommand{\ours}{\textsc{EnTraR}\xspace}   
\newcommand{\DevS}{Dev-S\xspace}
\newcommand{\DevC}{Dev-C\xspace}
\newcommand{\pp}{\,pp\xspace}

\newcommand{\pend}[1]{\textcolor{red}{#1}}

\newcommand{\vBaseRoneH}{20.23}  \newcommand{\vBaseRfiveH}{36.78} \newcommand{\vBaseRtenH}{44.30} \newcommand{\vBaseRhunH}{63.26}
\newcommand{\vBaseRoneS}{11.07}  \newcommand{\vBaseRfiveS}{24.45} \newcommand{\vBaseRtenS}{31.40} \newcommand{\vBaseRhunS}{47.97}
\newcommand{\vBaseVR}{42.47}     \newcommand{\vBaseVisH}{17.14}   \newcommand{\vBaseVisS}{9.21}
\newcommand{\vEnsRoneH}{22.83} \newcommand{\vEnsRfiveH}{39.72} \newcommand{\vEnsRtenH}{47.81} \newcommand{\vEnsRhunH}{66.69}
\newcommand{\vEnsRoneS}{12.76} \newcommand{\vEnsRfiveS}{27.64} \newcommand{\vEnsRtenS}{34.96} \newcommand{\vEnsRhunS}{51.69}
\newcommand{\vEnsVR}{45.56}    \newcommand{\vEnsVisH}{19.65}   \newcommand{\vEnsVisS}{10.63}
\newcommand{\vSelRoneH}{22.58} \newcommand{\vSelRfiveH}{39.52} \newcommand{\vSelRtenH}{46.87} \newcommand{\vSelRhunH}{63.26}
\newcommand{\vSelRoneS}{12.25} \newcommand{\vSelRfiveS}{27.00} \newcommand{\vSelRtenS}{34.05} \newcommand{\vSelRhunS}{47.97}
\newcommand{\vSelVR}{49.76}    \newcommand{\vSelVisH}{18.27}   \newcommand{\vSelVisS}{9.86}
\newcommand{\vFullRoneH}{24.36} \newcommand{\vFullRfiveH}{41.64} \newcommand{\vFullRtenH}{49.36} \newcommand{\vFullRhunH}{66.69}
\newcommand{\vFullRoneS}{13.55} \newcommand{\vFullRfiveS}{29.12} \newcommand{\vFullRtenS}{36.70} \newcommand{\vFullRhunS}{51.69}
\newcommand{\vFullVR}{51.49}    \newcommand{\vFullVisH}{19.82}   \newcommand{\vFullVisS}{10.77}
\newcommand{\dEnsH}{+2.60}  \newcommand{\ciEnsH}{[2.00, 3.17]}  \newcommand{\dEnsS}{+1.69}  \newcommand{\ciEnsS}{[1.22, 2.16]}
\newcommand{\dSelH}{+2.35}  \newcommand{\ciSelH}{[1.90, 2.76]}  \newcommand{\dSelS}{+1.18}  \newcommand{\ciSelS}{[0.86, 1.51]}
\newcommand{\dFullH}{+4.13} \newcommand{\ciFullH}{[3.48, 4.76]} \newcommand{\dFullS}{+2.48} \newcommand{\ciFullS}{[1.97, 2.99]}
\newcommand{\dVerH}{+1.53}  \newcommand{\ciVerH}{[1.07, 1.98]}  \newcommand{\dVerS}{+0.79}  \newcommand{\ciVerS}{[0.45, 1.13]}
\newcommand{\vVerSec}{0.48}

{
    \small
    \bibliographystyle{ieeenat_fullname}
    \bibliography{main}

@String(CVPR= {IEEE Conf. Comput. Vis. Pattern Recog.})

@String(ICCV= {Int. Conf. Comput. Vis.})

@String(CVPR  = {CVPR})

@String(ICCV  = {ICCV})

@inproceedings{bain2021frozen,
  author = {Max Bain and Arsha Nagrani and G{\"u}l Varol and Andrew Zisserman},
  title = {{Frozen in Time: A Joint Video and Image Encoder for End-to-End Retrieval}},
  booktitle = {Proceedings of the IEEE/CVF International Conference on Computer Vision},
  year = {2021}
}

@article{luo2022clip4clip,
  author = {Huaishao Luo and Lei Ji and Ming Zhong and Yang Chen and Wen Lei and Nan Duan and Tianrui Li},
  title = {{CLIP4Clip: An Empirical Study of CLIP for End to End Video Clip Retrieval and Captioning}},
  journal = {Neurocomputing},
  year = {2022},
  doi = {10.1016/j.neucom.2022.07.028}
}

@inproceedings{chen2020hgr,
  author = {Shizhe Chen and Yida Zhao and Qin Jin and Qi Wu},
  title = {{Fine-Grained Video-Text Retrieval with Hierarchical Graph Reasoning}},
  booktitle = {Proceedings of the IEEE/CVF Conference on Computer Vision and Pattern Recognition},
  year = {2020}
}

@inproceedings{reddy2025videocolbert,
  author = {Arun Reddy and Alexander Martin and Eugene Yang and Andrew Yates and Kate Sanders and Kenton Murray and Reno Kriz and Celso M. de Melo and Benjamin Van Durme and Rama Chellappa},
  title = {{Video-ColBERT: Contextualized Late Interaction for Text-to-Video Retrieval}},
  booktitle = {Proceedings of the IEEE/CVF Conference on Computer Vision and Pattern Recognition},
  year = {2025}
}

@inproceedings{dong2022prvr,
  author = {Jianfeng Dong and Xianke Chen and Minsong Zhang and Xun Yang and Shujie Chen and Xirong Li and Xun Wang},
  title = {{Partially Relevant Video Retrieval}},
  booktitle = {Proceedings of the ACM International Conference on Multimedia},
  year = {2022},
  doi = {10.1145/3503161.3547976}
}

@inproceedings{li2026dreamprvr,
  author = {Jun Li and Xuhang Lou and Jinpeng Wang and Yuting Wang and Yaowei Wang and Shu-Tao Xia and Bin Chen},
  title = {{Imagine Before Concentration: Diffusion-Guided Registers Enhance Partially Relevant Video Retrieval}},
  booktitle = {Proceedings of the IEEE/CVF Conference on Computer Vision and Pattern Recognition},
  year = {2026}
}

@inproceedings{gao2017tall,
  author = {Jiyang Gao and Chen Sun and Zhenheng Yang and Ram Nevatia},
  title = {{TALL: Temporal Activity Localization via Language Query}},
  booktitle = {Proceedings of the IEEE/CVF International Conference on Computer Vision},
  year = {2017}
}

@inproceedings{lei2021momentdetr,
  author = {Jie Lei and Tamara L. Berg and Mohit Bansal},
  title = {{Detecting Moments and Highlights in Videos via Natural Language Queries}},
  booktitle = {Advances in Neural Information Processing Systems},
  year = {2021}
}

@inproceedings{moon2023qddetr,
  author = {WonJun Moon and Sangeek Hyun and SangUk Park and Dongchan Park and Jae-Pil Heo},
  title = {{Query-Dependent Video Representation for Moment Retrieval and Highlight Detection}},
  booktitle = {Proceedings of the IEEE/CVF Conference on Computer Vision and Pattern Recognition},
  year = {2023}
}

@inproceedings{tang2025simdetr,
  author = {Jiajin Tang and Zhengxuan Wei and Yuchen Zhu and Cheng Shi and Guanbin Li and Liang Lin and Sibei Yang},
  title = {{Sim-DETR: Unlock DETR for Temporal Sentence Grounding}},
  booktitle = {Proceedings of the IEEE/CVF International Conference on Computer Vision},
  year = {2025}
}

@inproceedings{yang2023deco,
  author = {Lijin Yang and Quan Kong and Hsuan-Kung Yang and Wadim Kehl and Yoichi Sato and Norimasa Kobori},
  title = {{DeCo: Decomposition and Reconstruction for Compositional Temporal Grounding via Coarse-to-Fine Contrastive Ranking}},
  booktitle = {Proceedings of the IEEE/CVF Conference on Computer Vision and Pattern Recognition},
  year = {2023}
}

@inproceedings{an2026hieramamba,
  author = {Joungbin An and Kristen Grauman},
  title = {{HieraMamba: Video Temporal Grounding via Hierarchical Anchor-Mamba Pooling}},
  booktitle = {Proceedings of the IEEE/CVF Conference on Computer Vision and Pattern Recognition},
  year = {2026}
}

@article{escorcia2019moments,
  author = {Victor Escorcia and Mattia Soldan and Josef Sivic and Bernard Ghanem and Bryan Russell},
  title = {{Finding Moments in Video Collections Using Natural Language}},
  journal = {arXiv preprint arXiv:1907.12763},
  year = {2019}
}

@inproceedings{lei2020tvr,
  author = {Jie Lei and Licheng Yu and Tamara L. Berg and Mohit Bansal},
  title = {{TVR: A Large-Scale Dataset for Video-Subtitle Moment Retrieval}},
  booktitle = {European Conference on Computer Vision},
  year = {2020},
  doi = {10.1007/978-3-030-58589-1_27}
}

@inproceedings{zhang2021reloclnet,
  author = {Hao Zhang and Aixin Sun and Wei Jing and Guoshun Nan and Liangli Zhen and Joey Tianyi Zhou and Rick Siow Mong Goh},
  title = {{Video Corpus Moment Retrieval with Contrastive Learning}},
  booktitle = {Proceedings of the International ACM SIGIR Conference on Research and Development in Information Retrieval},
  year = {2021},
  doi = {10.1145/3404835.3462874}
}

@inproceedings{hou2021conquer,
  author = {Zhijian Hou and Chong-Wah Ngo and Wing Kwong Chan},
  title = {{CONQUER: Contextual Query-aware Ranking for Video Corpus Moment Retrieval}},
  booktitle = {Proceedings of the ACM International Conference on Multimedia},
  year = {2021}
}

@inproceedings{yoon2022squidnet,
  author = {Sunjae Yoon and Ji Woo Hong and Eunseop Yoon and Dahyun Kim and Junyeong Kim and Hee Suk Yoon and Chang D. Yoo},
  title = {{Selective Query-Guided Debiasing for Video Corpus Moment Retrieval}},
  booktitle = {European Conference on Computer Vision},
  year = {2022}
}

@inproceedings{hou2024eventformer,
  title     = {Event-aware Video Corpus Moment Retrieval},
  author    = {Hou, Danyang and Pang, Liang and Lan, Yanyan and Shen, Huawei and Cheng, Xueqi},
  booktitle = {European Conference on Information Retrieval (ECIR)},
  year      = {2026}
}

@inproceedings{chen2024verified,
  author = {Houlun Chen and Xin Wang and Hong Chen and Zeyang Zhang and Wei Feng and Bin Huang and Jia Jia and Wenwu Zhu},
  title = {{VERIFIED: A Video Corpus Moment Retrieval Benchmark for Fine-Grained Video Understanding}},
  booktitle = {Advances in Neural Information Processing Systems},
  year = {2024}
}

@inproceedings{pan2025refcap,
  author = {Yi Pan and Yujia Zhang and Michael Kampffmeyer and Xiaoguang Zhao},
  title = {{RefCap: Zero-shot Video Corpus Moment Retrieval Based on Refined Dense Video Captioning}},
  booktitle = {IEEE International Conference on Acoustics, Speech and Signal Processing},
  year = {2025},
  doi = {10.1109/ICASSP49660.2025.10888164}
}

@article{zhao2026scac,
  author = {Jialong Zhao and Huafeng Li and Yafei Zhang and Wen Wang and Changchun Hua},
  title = {{Training-Free Video Corpus Moment Retrieval via Synergistic Collaboration and Adaptive Calibration}},
  journal = {IEEE Transactions on Image Processing},
  year = {2026},
  doi = {10.1109/TIP.2026.3723243}
}

@article{yang2026demul,
  author = {Yong Yang and Meng Liu and Xuemeng Song and Na Zheng and Ke Lu and Weili Guan},
  title = {{Video Corpus Moment Retrieval via Decoupled Multimodal Modeling and Unified Localization}},
  journal = {IEEE Transactions on Circuits and Systems for Video Technology},
  year = {2026},
  doi = {10.1109/TCSVT.2026.3705906}
}

@article{sun2026genspan,
  author = {Yunzhuo Sun and Xinyue Liu and Yanyang Li and Nanding Wu and Linlin Zong and Xianchao Zhang and Wenxin Liang},
  title = {{GenSpan: Generation-Calibrated Motion Span Priors for Multi-Verb Video Corpus Moment Retrieval}},
  journal = {arXiv preprint arXiv:2603.22121v2},
  year = {2026}
}

@inproceedings{lee2026videosearch,
  author = {Seohyun Lee and Seoung Choi and Dohwan Ko and Jongha Kim and Hyunwoo J. Kim},
  title = {{VideoSearch-R1: Iterative Video Retrieval and Reasoning via Soft Query Refinement}},
  booktitle = {European Conference on Computer Vision},
  year = {2026}
}

@inproceedings{li2020hero,
  title     = {{HERO}: Hierarchical Encoder for Video+Language Omni-representation Pre-training},
  author    = {Li, Linjie and Chen, Yen-Chun and Cheng, Yu and Gan, Zhe and Yu, Licheng and Liu, Jingjing},
  booktitle = {Conference on Empirical Methods in Natural Language Processing (EMNLP)},
  year      = {2020}
}

@article{zhang2020hammer,
  title   = {A Hierarchical Multi-Modal Encoder for Moment Localization in Video Corpus},
  author  = {Zhang, Bowen and Hu, Hexiang and Lee, Joonseok and Zhao, Ming and Chammas, Sheide and Jain, Vihan and Ie, Eugene and Sha, Fei},
  journal = {arXiv preprint arXiv:2011.09046},
  year    = {2020}
}

@article{hou2023minute,
  title   = {Multi-video Moment Ranking with Multimodal Clue},
  author  = {Hou, Danyang and Pang, Liang and Lan, Yanyan and Shen, Huawei and Cheng, Xueqi},
  journal = {arXiv preprint arXiv:2301.13606},
  year    = {2023}
}

@inproceedings{hou2024prem,
  title     = {Improving Video Corpus Moment Retrieval with Partial Relevance Enhancement},
  author    = {Hou, Danyang and Pang, Liang and Shen, Huawei and Cheng, Xueqi},
  booktitle = {ACM International Conference on Multimedia Retrieval (ICMR)},
  year      = {2024}
}

@inproceedings{radford2021clip,
  title     = {Learning Transferable Visual Models From Natural Language Supervision},
  author    = {Radford, Alec and Kim, Jong Wook and Hallacy, Chris and Ramesh, Aditya and Goh, Gabriel and Agarwal, Sandhini and Sastry, Girish and Askell, Amanda and Mishkin, Pamela and Clark, Jack and Krueger, Gretchen and Sutskever, Ilya},
  booktitle = {International Conference on Machine Learning (ICML)},
  year      = {2021}
}

@inproceedings{he2016resnet,
  title     = {Deep Residual Learning for Image Recognition},
  author    = {He, Kaiming and Zhang, Xiangyu and Ren, Shaoqing and Sun, Jian},
  booktitle = {IEEE Conference on Computer Vision and Pattern Recognition (CVPR)},
  year      = {2016}
}

@inproceedings{feichtenhofer2019slowfast,
  title     = {{SlowFast} Networks for Video Recognition},
  author    = {Feichtenhofer, Christoph and Fan, Haoqi and Malik, Jitendra and He, Kaiming},
  booktitle = {IEEE/CVF International Conference on Computer Vision (ICCV)},
  year      = {2019}
}

@inproceedings{xiao2024cpack,
  title     = {{C-Pack}: Packed Resources For General {Chinese} Embeddings},
  author    = {Xiao, Shitao and Liu, Zheng and Zhang, Peitian and Muennighoff, Niklas and Lian, Defu and Nie, Jian-Yun},
  booktitle = {ACM SIGIR Conference on Research and Development in Information Retrieval},
  year      = {2024}
}

@inproceedings{qi2020stanza,
  title     = {Stanza: A {Python} Natural Language Processing Toolkit for Many Human Languages},
  author    = {Qi, Peng and Zhang, Yuhao and Zhang, Yuhui and Bolton, Jason and Manning, Christopher D.},
  booktitle = {Annual Meeting of the Association for Computational Linguistics: System Demonstrations},
  year      = {2020}
}

@misc{qwen35,
  title        = {{Qwen3.5}},
  author       = {{Qwen Team}},
  howpublished = {Model release, \url{https://huggingface.co/Qwen/Qwen3.5-27B}},
  year         = {2026}
}

@inproceedings{kwon2023vllm,
  title     = {Efficient Memory Management for Large Language Model Serving with {PagedAttention}},
  author    = {Kwon, Woosuk and Li, Zhuohan and Zhuang, Siyuan and Sheng, Ying and Zheng, Lianmin and Yu, Cody Hao and Gonzalez, Joseph E. and Zhang, Hao and Stoica, Ion},
  booktitle = {ACM Symposium on Operating Systems Principles (SOSP)},
  year      = {2023}
}

@article{hinton2002poe,
  title   = {Training Products of Experts by Minimizing Contrastive Divergence},
  author  = {Hinton, Geoffrey E.},
  journal = {Neural Computation},
  volume  = {14},
  number  = {8},
  pages   = {1771--1800},
  year    = {2002}
}

@book{efron1993bootstrap,
  title     = {An Introduction to the Bootstrap},
  author    = {Efron, Bradley and Tibshirani, Robert J.},
  publisher = {Chapman \& Hall},
  year      = {1993}
}
}


\end{document}